\documentclass[conference]{IEEEtran}
\IEEEoverridecommandlockouts
\usepackage{amsmath,amsfonts}
\usepackage{algorithmic}
\usepackage{algorithm}
\usepackage{array}
\usepackage[caption=false,font=normalsize,labelfont=sf,textfont=sf]{subfig}
\usepackage{booktabs}
\usepackage{textcomp}
\usepackage{stfloats}
\usepackage{url}
\usepackage{verbatim}
\usepackage{graphicx}
\usepackage{cite}
\usepackage{pifont}
\usepackage{fontawesome}
\usepackage{multirow}
\usepackage{caption}
\newcommand{\hupar}[1]{\vspace{12pt} \noindent \textbf{#1}\quad}

\renewcommand{\algorithmicrequire}{\textbf{Input:}}
\renewcommand{\algorithmicensure}{\textbf{Output:}}

\usepackage{xcolor}
\def\BibTeX{{\rm B\kern-.05em{\sc i\kern-.025em b}\kern-.08em
    T\kern-.1667em\lower.7ex\hbox{E}\kern-.125emX}}
\begin{document}

\title{SCSim: A Realistic Spike Cameras Simulator}

\author{Liwen Hu\textsuperscript{\rm 1},
Lei Ma\textsuperscript{\rm 1,2*},
Yijia Guo\textsuperscript{\rm 1}, 
Tiejun Huang\textsuperscript{\rm 1}\\
\textsuperscript{\rm 1}NERCVT, School of Computer Science, Peking University\\
\textsuperscript{\rm 2}College of Future Technology, Peking University\\
}

\twocolumn[{%
\renewcommand\twocolumn[1][]{#1}%
\maketitle
\begin{center}
    \centering
    \captionsetup{type=figure}
    \includegraphics[width=1.0\textwidth]{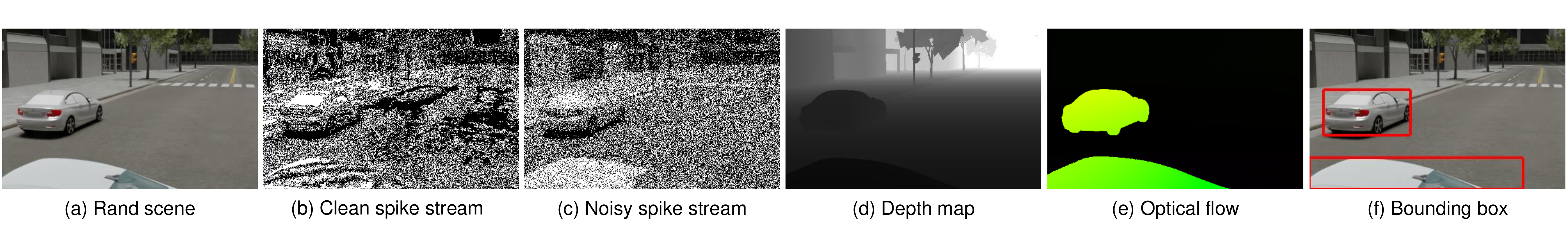}
    \captionof{figure}{An example of the output of proposed simulator, SCSim, in a rand scene.}\label{label}
\end{center}%
}]


\begin{abstract}
Spike cameras, with their exceptional temporal resolution, are revolutionizing high-speed visual applications. Large-scale synthetic datasets have significantly accelerated the development of these cameras, particularly in reconstruction and optical flow. However, current synthetic datasets for spike cameras lack sophistication. Addressing this gap, we introduce SCSim, a novel and more realistic spike camera simulator with a comprehensive noise model. SCSim is adept at autonomously generating driving scenarios and synthesizing corresponding spike streams. To enhance the fidelity of these streams, we've developed a comprehensive noise model tailored to the unique circuitry of spike cameras. Our evaluations demonstrate that SCSim outperforms existing simulation methods in generating authentic spike streams. Crucially, SCSim simplifies the creation of datasets, thereby greatly advancing spike-based visual tasks like reconstruction. Our project refers to https://github.com/Acnext/SCSim.
\end{abstract}

\footnotetext{* Corresponding author. \\\indent
This work was supported by National Science and Technology Major Project (2022ZD0116305).}



\section{Introduction}

Neuromorphic cameras, including event cameras \cite{dvs1, Weng_2023_CVPR} and spike cameras \cite{spikecamera}, have the advantages of high temporal resolution and low redundancy. Unlike the differential sampling mechanism in event cameras, spike cameras record absolute light intensity by releasing asynchronous spikes. Therefore, spike cameras can not only capture dynamic scenes but also texture details.
It has shown enormous potential for high-speed visual tasks such as reconstruction \cite{rec5}, optical flow estimation \cite{2022scflow}, and depth estimation \cite{2022spkTransformer}. However, the development of spike cameras is still in its early stages and the lack of large-scale datasets has greatly limited it.

Recently, a meaningful spike camera simulator, SPCS \cite{2022scflow}, is proposed. It can construct spike stream datasets as two steps:
Step.1. Generating high frame rate image sequences for high-speed scenes.
Step.2. Convert the image sequence to a spike stream based on spike camera model. 
However, due to the neglect of scene quality and noise models, generated spike streams are not realistic enough.

To reduce the gap between the generated spike streams and the real spike streams, this work proposes a more realistic spike camera simulator. It improves step.1 and step.2 respectively.  For step.1, SPCS \cite{2022scflow} provides a rand scenes function to generate image sequences of rand high-speed scenes. However, rand high-speed scenes describe multiple floating objects which do not exist in the real world. To this end, we design a new rand scenes function where random drive scenes can be conveniently generated as shown in Fig.~\ref{label}. In addition, to enrich our 
rand scenes, we also provide a variety of lighting conditions.  
For step.2, SPCS \cite{2022scflow} lacks the spike camera noise model which cannot accurately simulate the working process of the spike camera. Although NeuSpike \cite{rec4} and SpikingSIM \cite{zhao2022spikingsim} consider the noise in the spike camera, their noise model is relatively simple where the dark current is considered as the main noise source.
We delve deeper into the distinct noise mechanism in a spike camera. Further, we propose a realistic simulation method for spike cameras where its circuit-level noise is modeled. By building the relationship between spike stream and luminance intensity, we
propose the spike-based noise evaluation equation
(SNEE). Finally, we set up the noise measurement experiment and evaluate the statistics of the noise variable based on SNEE. 

Experiments show that SCSim can generate more realistic spike streams than other methods. In addition, a random high-speed driving dataset, RHDD, is generated by SCSim. We use RHDD to finetune the state-of-the-art reconstruction method, WGSE \cite{rec5}. The results on real data show that the reconstructed images from WGSE (finetune) have a lower noise level.  Our main contributions are summarized as follows:
\begin{itemize}
\item[$\bullet$] \textbf{A spike camera simulation method:} We analyze the basic principles of spike camera circuit implementation in detail and model its distinct noise.

\item[$\bullet$] \textbf{A measurement method for spike camera noise:} We propose the Spike-based Noise Evaluation Equation (SNEE) to establish the relationship between noise and spike stream, and we collect real spike stream and use SNEE to evaluate the statistics of noise.

\item[$\bullet$] \textbf{Rich auxiliary functions:} Our simulator, SCSim, provides random driving scene and label generation. It allows us to easily build spike stream datasets.


\item[$\bullet$] \textbf{Effectiveness evaluation of simulator} 
We find that SCSim can generate more realistic data than other methods. Besides, datasets from SCSim can improve the performance of the state-of-the-art reconstruction method.

%

\end{itemize}





\section{Related Work}

\subsection{Spike Stream Reconstruction}
The spike camera shows potential in high-speed visual tasks. Spike stream reconstruction is always a fundamental task. Based on the statistical characteristics of spike stream, TFI and TFP \cite{spikecamera} first reconstruct high-speed scenes. Spiking neural networks \cite{rec1, rec2} and convolutional neural networks \cite{rec3, rec4} are respectively used  to reconstruct high-speed images from a spike stream, which greatly improves the reconstruction quality. WGSE \cite{rec5} constructs a  wavelet representation to better deal with noise in spike streams.

\subsection{Spike Camera Simulation}

S2I \cite{rec3} generate the spike stream reconstruction training set based on spike camera simulation. Specifically, it first generates high frame rate image sequences from existing datasets based on video interpolation methods. Then, the image sequence is converted into spike streams.
To obtain high-speed image sequences more freely, the spike camera simulator (SPCS) \cite{2022scflow} is proposed and it combines simulation function and rendering engine tightly. 
However, the above method does not model the noise in the spike camera. NeuSpike \cite{rec4} and SpikingSIM \cite{zhao2022spikingsim} add dark current noise to the ideal spike camera model for more accurate generation.





\section{Spike Camera Simulator}

\begin{figure*}[tp]
\includegraphics[width=\linewidth]{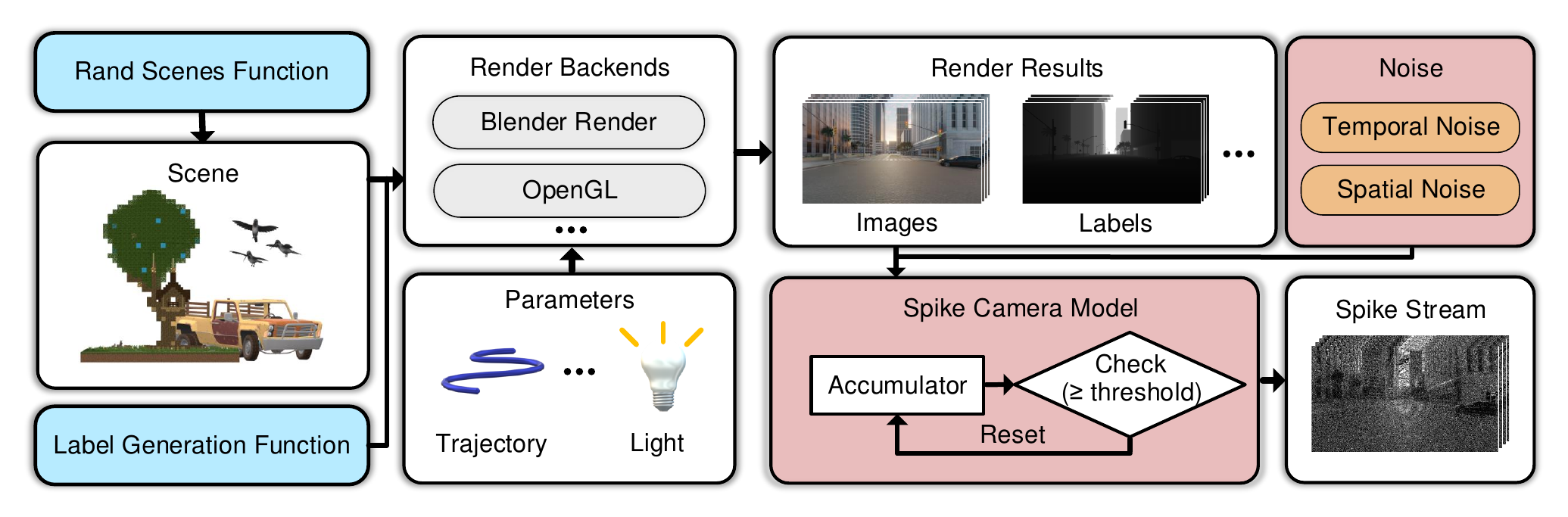}
\centering
\caption{Framework of spike camera simulator, SCSim. First, according to graphic scenes and camera parameters, image sequences, and vision task labels are generated by render backends. Then we convert the image sequences into spike streams based on our spike camera model. The blue (red) boxes denote the auxiliary functions (the spike camera model) in SCSim.}\label{framework}
\end{figure*}


\subsection{Overview}
As shown in Fig.~\ref{framework}, we demonstrate the framework of SCSim where we convert rendered images in high-speed scenes into spike streams. 
By using rand scenes function and label generation function, we can build high-speed scenes quickly before rendering and obtain  labels for various visual tasks during the rendering process. Finally, we can generate more various datasets based on our spike camera model.
\vspace{-6pt}
\subsection{Spike Camera Model}
Based on the circuit principle of the spike camera, we model the joint interference of various noises for the generation of spike streams. We start by introducing a noise-free spike camera model. And then we propose a practical model along with the noises in the spike camera.
\subsubsection{Idea Spike Generation}
The spike camera mimicking the retina fovea consists of an array of $H \times W$ pixels and can report per-pixel luminance intensity by firing asynchronous spikes. Specifically, as shown in Fig.~\ref{spike_camera_model}, each pixel on the spike camera sensor accumulates incoming light independently and persistently. At the time $t$, for pixel  $(x, y)$, if the accumulated brightness arrives at a fixed threshold  $\phi$ (as (1)), a spike is fired and then the accumulated brightness can be reset to 0.
   \begin{eqnarray}
    {A}(x, y, t) = \int_{t_{x, y}^{\rm pre}}^{t} {I_{in}}(x, y, \tau) d\tau\ \geq \phi, 
    \end{eqnarray}
    where  $x, y\in \mathbb{Z}, x \leq H, y \leq W$, $  {A}(x, y, t)$  is the accumulated brightness at time  $t$,  ${I_{in}}(x, y, \tau)$ is the input and $t_{x, y}^{\rm pre}$ expresses the last time when a spike is fired at pixel  $(x, y)$  before time  $t$. If  $t$   is the first time to send a spike, then  $t_{x, y}^{\rm pre}$  is set as 0. In fact, due to the limitations of circuit technology, the spike reading times are quantified. Hence, asynchronous spikes are read out synchronously. Specifically, all pixels periodically judge the spike flag at time $n\delta t, n \in \mathbb{Z}$, where $\delta t$  is a short interval of microseconds.     
    Therefore, the output of all pixels forms a  $H \times W$   binary spike frame. As time goes on, the camera would produce a sequence of spike frames, i.e., a  $H \times W \times N$  
binary spike stream and can be mathematically defined as, 
   \begin{eqnarray}
    \begin{aligned}
    &  {S}(x, y, n\delta t) =
    \\&
    \begin{cases}
    1 &\mbox{ if  $\exists t \in \left((n - 1)\delta t, n\delta t\right]$, s.t.  $  {A}(x, y, t) \geq \phi$   }, \\
    0 &\mbox{ if  $\forall t \in \left((n - 1)\delta t, n\delta t\right]$, $  {A}(x, y, t) < \phi$  } \\
    \end{cases} 
    \end{aligned}
    \end{eqnarray}

\subsubsection{Spike Generation with Noises} 
\begin{figure}[tbp]
\includegraphics[width=\linewidth]{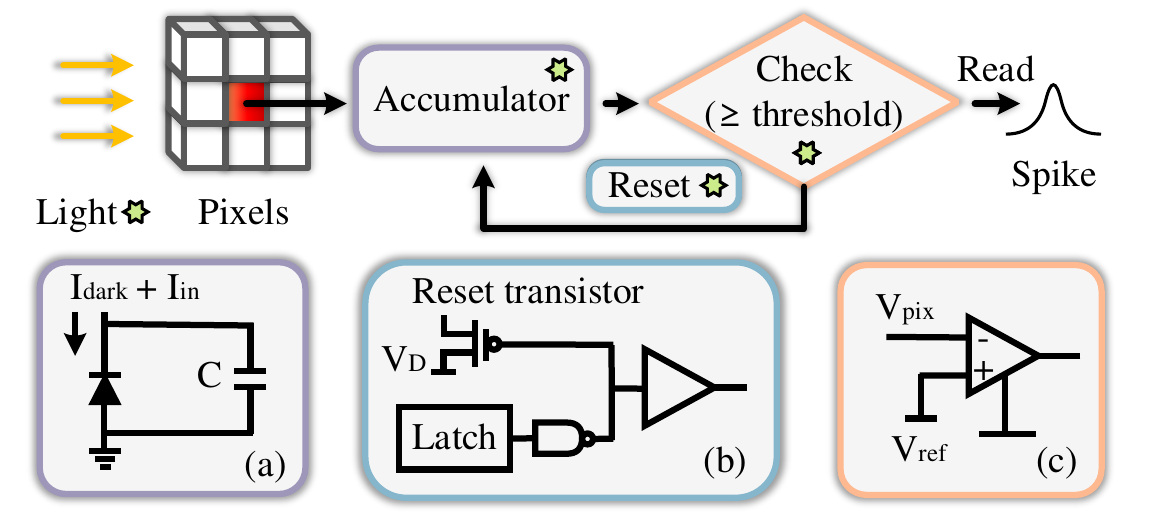}
\centering
\caption{The principle and circuit of spike camera. (a-c) is the working principle of the accumulator, reset model, and check mechanism in a pixel circuit of spike camera. ``Star" means that we have considered the corresponding noise.}\label{spike_camera_model}
\end{figure}

We systematically introduce the temporal noise and spatial noise in the spike camera according to its unique circuit as shown in Fig.~\ref{spike_camera_model}(a-c).

\hupar{Temporal Noise} Temporal noise is a random variation in the signal that fluctuates over time. The temporal noise in the spike camera mainly includes shot noise and thermal noise. The shot noise originates from randomness
caused by photon reception and, for pixel $(x, y)$, the probability that $n$ photons are received between time $t$ and $t + \delta t$ is given by the Poisson probability distribution, i.e., 
\begin{equation}
P({ph}(x, y, t) = n) = \frac{\mu_{  {ph}(x, y, t)}}{n!e^{\mu_{  {ph}(x, y, t)}}}
\end{equation}
where $n \in \mathbb{N}$, $  {ph}(x, y, t)$ is a random variable representing the number of received photons from $t$ to $t + \delta t$, and $\mu_{  {ph}(x, y, t)}$ is the expectation of $  {ph}(x, y, t)$.
The random number of photons can affect the luminance intensity. Since $\delta t$ is enough small, we consider the luminance intensity at time $t$ to be proportional to the number of photons between time $t$ and $t + \delta t$, i.e., $  {L}(x, y, t) \propto   {ph}(x, y, t)$.
Different from the previous simulator \cite{2022scflow} which gets the input current (see (1)) referring to the ideal luminance intensity $\mu_{{L}(x, y, t)}$, we first sample $  {L}(x, y, t)$ at time $t$ during simulation, i.e.,
\begin{equation}
  {L}(x, y, t) = \mu_{  {L}(x, y, t)}\dfrac{  {ph}(x, y, t)}{\mu_{  {ph}(x, y, t)}}.
\end{equation}
Further, the input current $ {I_{in}}(x, y, t)$ at time $t$ can be expressed as $\alpha{L}(x, y, \tau)$ where $\alpha$ is the photoelectric conversion rate and it can be estimated through the dynamic range of spike camera. 
The threshold $\phi$ (as (1)) in the ideal spike camera model also fluctuates over time. We start by writing the ideal threshold in the form of the circuit, i.e,
\begin{equation}
\phi = CV_{d} = C(V_{D} - V_{ref})
\end{equation} The voltage fluctuation in the camera can be caused by the reset transistor, which is affected by temperature
where $V_{D}$ is the reset voltage and $V_{ref}$ is the reference voltage. As shown in Fig.~\ref{spike_camera_model}(b), the reset transistor affected by temperature can cause voltage fluctuation. We use the random variable ${V}^{T_0}(x,y,t)$ to describe fluctuating voltage in pixel $(x, y)$ at time $t$ and ${V}^{T_0}(x,y,t)$ can be considered to obey Gaussian distribution, i.e., 
\begin{equation}
  {V}^{T_0}(x,y,t) \sim N(0, (\sigma^{T_0})^2), \quad  \sigma^{T_0} = \sqrt{\frac{kT_0}{C}},
\end{equation}
where $\sigma^{T_0}$ is the standard deviation of $  {V}^{T_0}(x, y, t)$, $k$ is Boltzmann constant and $T_0$ is absolute temperature.

\hupar{Spatial Noise}
Spatial noise, i.e., fixed-pattern noise, is a random variation in the signal that has nothing to do with time. For the spatial noise in the spike camera, we mainly consider dark current, capacitor mismatch, and bias voltage. The dark current is from photodiodes and it can generate extra electricity as accumulation in spike camera. The dark current $I_{\rm {dark}}(x, y)$ in pixel $(x, y)$ obeys Gaussian distribution as,
\begin{equation}
I_{\rm {dark}}(x, y) \sim N(\mu_{\rm {dark}}, (\sigma_{\rm {dark}}^{S})^2),
\end{equation}
where  $\mu_{\rm {dark}}$ and $\sigma_{\rm {dark}}^{S}$ contribute to the expectation and variance of $I_{\rm {dark}}(x, y)$. Besides, the capacitance difference in each pixel obeys Gaussian distribution and can be written as, 
\begin{equation}
C^{S}(x,y) \sim N(0, (\sigma_{C}^{S})^2),
\end{equation}
where $C^{S}(x,y)$ is the random variable describing the capacitance nonuniformity at pixel $(x, y)$, $\sigma_{C}^{S}$ is the standard deviation of $C^{S}(x,y)$.
The bias voltage $V^{S}(x,y)$ mainly comes from reset voltage nonuniformity in the check module as shown in Fig.~\ref{spike_camera_model}(c). We can assume the bias voltage $V^{S}(x,y)$ at pixel $(x, y)$ obeys Gaussian distribution, i.e.,
\begin{equation}
V^{S}(x,y) \sim N(0, (\sigma_{V}^{S})^2),
\end{equation}
where  $\sigma_{V}^{S}$ is the standard deviation of $V^{S}(x,y)$.

\hupar{The Sampling Process with Noise}
We can add the above noise model to the generation of the spike stream. At time $t$, for pixel $(x, y)$, the accumulated brightness $  {A}(x, y, t)$ in (1) can be rewritten as,
\begin{equation}
\int_{t_{x, y}^{\rm pre}}^t \!\!\!\!\!\! \alpha {\color{red} L(x, y, \tau)} + {\color{red} I_{\rm {dark}}(x, y)} d\tau \geq \phi(x, y, t),
\end{equation}
where $\phi(x, y, t)$ is the threshold affected by noise.
Further, $\phi(x, y, t)$ can be expressed as,
\begin{equation}
\begin{aligned}
(C + {\color{blue} C^{S}(x, y)})(V_{d} + {\color{blue} V^{T_0}(x, y, t)} + {\color{blue} V^{S}(x, y)}).
\end{aligned}
\end{equation}
The spike stream $  {S}(x, y, n\delta t)$ in (2) can be rewritten as,
   \begin{equation}
    \begin{aligned}
    \begin{cases}
    1 \!\!\!\!&\mbox{ if  $\exists t \in \left((n - 1)\delta t, n\delta t\right]$, s.t.  $  {A}(x,y, t) \geq \phi(x,y,t)$}, \\
    0 \!\!\!\!&\mbox{ if  $\forall t \in \left((n - 1)\delta t, n\delta t\right]$, $  {A}(x,y, t) < \phi(x,y,t)$}.  \\
    \end{cases}
    \end{aligned}
    \end{equation}
Furthermore, we can simulate the sampling process in spike camera as Algorithm.~\ref{sim_alg}. 


\begin{algorithm}[ht]
\caption{Simulation of spike camera.} 
 \algorithmicrequire{ Idea luminance intensity sequences $\{i,x,y  \in \hspace*{0.02in} \mathbb{N^*} , i \leq T, x\leq H, y \leq W|\mu_{{L}(x, y, i\delta t)}\}$ and noise intensity \hspace*{0.02in} $\sigma_{C}^S$, $\sigma_{V}^{S}$, $\mu_{dark}$, $\sigma_{dark}^{S}$, $\mu_{\alpha}$, $\sigma_{\alpha}^{S}$, $\sigma^{T_0}$.}

 \algorithmicensure{ Noisy spike stream $\mathbf{S}_n^{0, T}$.}

\begin{algorithmic}[1]
    \STATE Sample spatial state, $\mathbf{C}^S$, $\mathbf{V}^S$, $\boldsymbol{\alpha}$ and $\mathbf{I}_{dark}$.
    \STATE Initialize accumlation, $\mathbf{A} = 0$, $\mathbf{S}_n^{0, T} = \mathbf{0}$.
    \FOR {each timestamp $i\delta t$}
        \STATE Sample temporal state, $\mathbf{V}^{T_0}$ and $\mathbf{L}(:, :, i\delta t)$ .
        \STATE Accumlate, $\mathbf{A} =\mathbf{A} + \boldsymbol{\alpha}\otimes \mathbf{L}(:, :, i\delta t)+\mathbf{I}_{dark}$. 
    \STATE Check, $[\mathbf{x}, \mathbf{y}]$ = $[\mathbf{A} \geq (\mathbf{C} + \mathbf{C}^{S})\otimes(\mathbf{V}_{d} + \mathbf{V}^{T_0} + \mathbf{V}^{S})]$.
        \STATE Fire spikes, $\mathbf{S}_n^{0, T}(\mathbf{x}, \mathbf{y}, i\delta t) = 1$.
        \STATE Reset, $ \mathbf{A} =  \mathbf{A} - (\mathbf{C} + \mathbf{C}^{S})\otimes(\mathbf{V}_{d} + \mathbf{V}^{T_0} + \mathbf{V}^{S})$.
    \ENDFOR
\STATE Write noisy spike stream $\mathbf{S}_n^{0, T}$.
\end{algorithmic}\label{sim_alg}
\end{algorithm}

\subsection{Auxiliary Functions}
We introduce two main auxiliary functions in SCSim that can be used to generate large-scale spike camera datasets.

\subsubsection{Rand Scenes Function}

\begin{figure}[htbp]
\includegraphics[width=\linewidth]{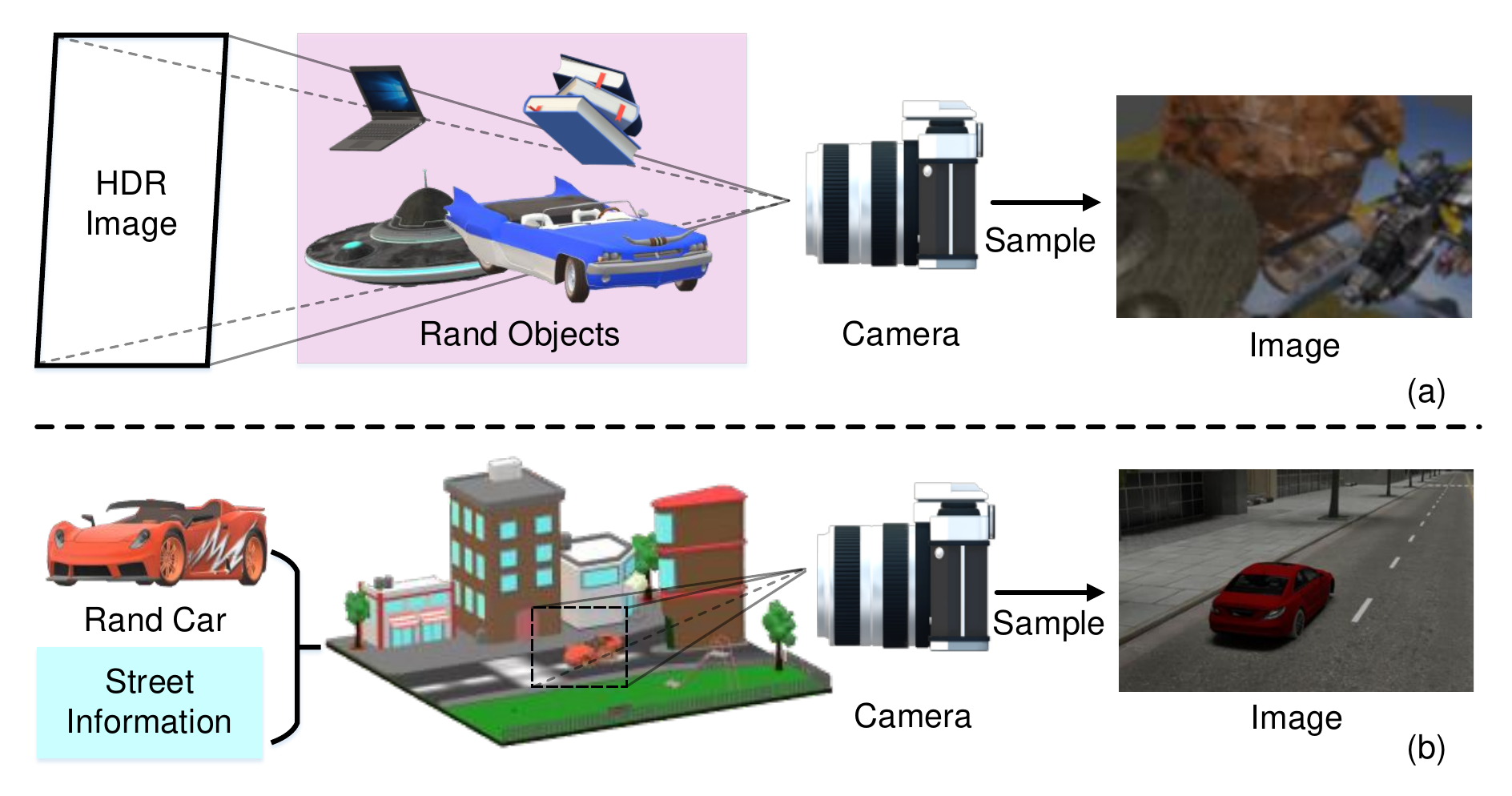}
\centering
\caption{The difference of rand scenes function between SPCS\cite{2022scflow} and our SCSim. (a) Rand scenes function in SPCS where rand objects move and the background is a static HDR image. (b) Rand scenes function in SCSim. By combining rand cars, street information, and city models, it can randomly generate automatic driving scenes.}\label{randsences}
\end{figure}

The spike camera simulator, SPCS, in \cite{2022scflow} also provides a random scene function, which can generate random moving objects. However, as shown in Fig.~\ref{randsences}(a), it has two obvious disadvantages: 

(a). The random scenes and the real scenes have a large gap: this gap can make generated training sets different from the real dataset. 

(b). Using HDR images as backgrounds can result in loss of physical properties: Backgrounds in random scenes can generate inaccurate label information, such as depth estimation and optical flow estimation. For depth estimation tasks, the depth of pixels in the background is always consistent. For optical flow estimation tasks, the motion of background pixels is similar. Naturally, using these data as a training dataset can reduce the performance of networks.


The random scene function in SCSim solves the above problems. As shown in Fig.~\ref{randsences}(b), our random scene function can generate an automatic driving scene according to three parts of cars, street information, and city model which is closer to real words than SPCS. More details are in the appendix. 


\subsubsection{Label Generation Function}
Our SCSim provides label generation functions for various visual tasks, e.g., optical flow and depth. Label-generation functions can be divided into two categories. One is the label generation functions supported by the rendering engine, namely optical flow and depth. We can directly call these functions. Another label generation function is not provided in the rendering engine, e.g., the bounding box of cars. Therefore, we have implemented the calculation and output of the labels. We use the rand scenes function and label generation function to generate spike stream and corresponding visual labels in autonomous driving scenarios.


\section{Model Calibration}
 To measure statistics of noise i.e., $\mathbf{C}^S$, $\mathbf{V}^S$ and $\mathbf{I}_{dark}$, a reasonable experiment is designed. We start by proposing a spike-based noise evaluation equation (SNEE) based on the relationship between noise and spike stream. Then, we build a measurement setup and sample real data. Using the sampled data and the SNEE equation, we complete model calibration, i.e., estimating the above noise variable. 

\subsection{Spike-based Noise Evaluation Equation}
A spike fired by pixel $(x, y)$ in time $t$ means that the accumulation at the pixel arrives at threshold $\phi(x, y, t)$. Hence, if accumulation would not be reset, the total accumulation at the pixel $(x, y)$ in time $n\delta t$ can be estimated as, 
\begin{equation}
\sum_{S(x, y, i\delta t) = 1}^{i}\!\!\!\!\!\!\!\! \phi(x, y, t_i),
\end{equation}
where $i \in \mathbb{Z} \cap [1, n]  \; s. t.  \; S(x, y, i\delta t) = 1$, $t_i$ express the time when the  i-th spike is fired and  $t_i \in ((i - 1)\delta t, i\delta t]$. According to (11), the total accumulation at pixel $(x, y)$ in time $n\delta t$  also can be expressed as,
\begin{equation}
\int_{0}^{n\delta t}\!\!\! \alpha L(x, y, \tau) + I_{\rm {dark}}(x, y) d\tau.
\end{equation}
Further, we have the spike-based noise evaluation equation (SNEE), i.e.,
\begin{equation}
\begin{aligned}
&\sum_{S(x, y, i\delta t) = 1}^i\!\!\!\!\!\!\!\! \phi(x, y, t_i)\\
&=\!\!\!\!\!\!\! \sum_{S(x, y, i\delta t) = 1}^i\!\!\!\!\!\!\!\! (C + C^{S}(x, y))(V_{d} + V^{T_0}(x, y, t_i) + V^{S}(x, y)) \\
&=\int_{0}^{n\delta t}\!\!\! \alpha L(x, y, \tau) + I_{\rm {dark}}(x, y) d\tau
\end{aligned}
\end{equation}
We can set experimental scenes to control $L(x, y, \tau)$ and eliminate temporal noise by extending sampling time $n\delta t$ (see appendix). When an experimental scene is static and sampling time $n\delta t$ is long enough, (15) can be simplified as,
\begin{equation}
\begin{aligned}
&\sum_{S_k(x, y, i\delta t) = 1}^i\!\!\!\!\!\!\!\! (C + C^{S}(x, y))(V_{d} + V^{S}(x, y)) \\
&=\int_{0}^{n\delta t}\!\!\! \alpha \mu_{k} + I_{\rm {dark}}(x, y) d\tau
\end{aligned}\label{eq_snee}
\end{equation}
where $k \in \mathbb{Z}$, $\mu_{k}$ is the ideal luminance intensity in the k-th static scene and $\mathbf{S}_k(x, y, t)$ is the spike stream captured from k-th static scene.

\begin{figure}[t]
\includegraphics[width=\linewidth]{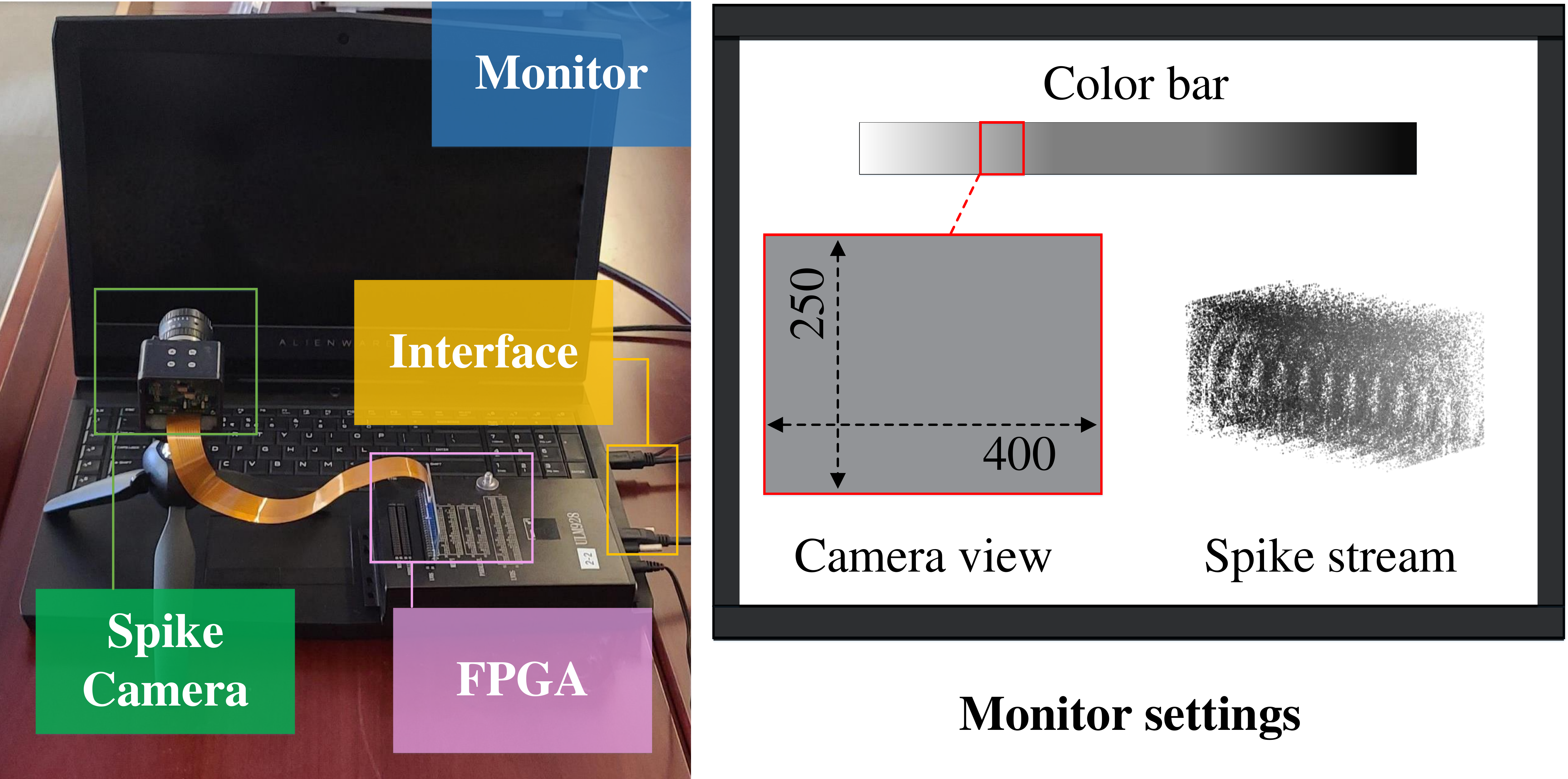}
\centering
\caption{The spike camera shoots a computer monitor under different grayscale backgrounds.}\label{set}
\end{figure}

\subsection{Noise Measurement Setup}
As shown in Fig.~\ref{set}, we built the experimental scenes where the spike camera shots computer monitor under 25 different grayscale backgrounds, i.e., $\{Gray_{k} = 0.1k|k \in [0, 24] \cap \mathcal{Z}\}$, respectively. Hence, we can estimate light intensity in the scenes as $(Gray_{k})*L_{monitor}$.
Further, accroding to (\ref{eq_snee}), we can get a system of equations about $I_{\rm {dark}}(x, y)$, $C^S(x,y)$ and $V^S(x, y)$ for each pixel $(x, y)$. We can solve the above equations by optimization methods. Specifically, in the k-th experimental scene, for each pixel $(x, y)$, we write the left side of (\ref{eq_snee}) as $Eql_k^{(x, y)}$ and the right side of (\ref{eq_snee}) as $Eqr_k^{(x, y)}$ where we omit the variables, $C^S, V^S, \alpha$ and $I_{dark}$ for simplicity.  Then, we have equations $\{k \in \mathbb{Z} | Eql_k^{(x, y)} - Eqr_k^{(x, y)} = 0\}$. Due to the existence of measurement error, we cannot directly get the exact solution of the equations. Therefore, we regard the solution as an optimization problem, i.e.,
\begin{equation}
\min_{C^S, V^S, \alpha, I_{dark}} \sum^k|Eql_k^{(x, y)} - Eqr_k^{(x, y)}|
\end{equation}
where $|\cdot|$ is the operation of absolute value and the interior-point method is used to solve it.
Finally, the statics of noise can be evaluated from $I_{\rm {dark}}(x, y)$, $C^S(x,y)$ and $V^S(x, y)$. Related details are in the appendix.

\begin{figure*}[t]
\includegraphics[width=\linewidth]{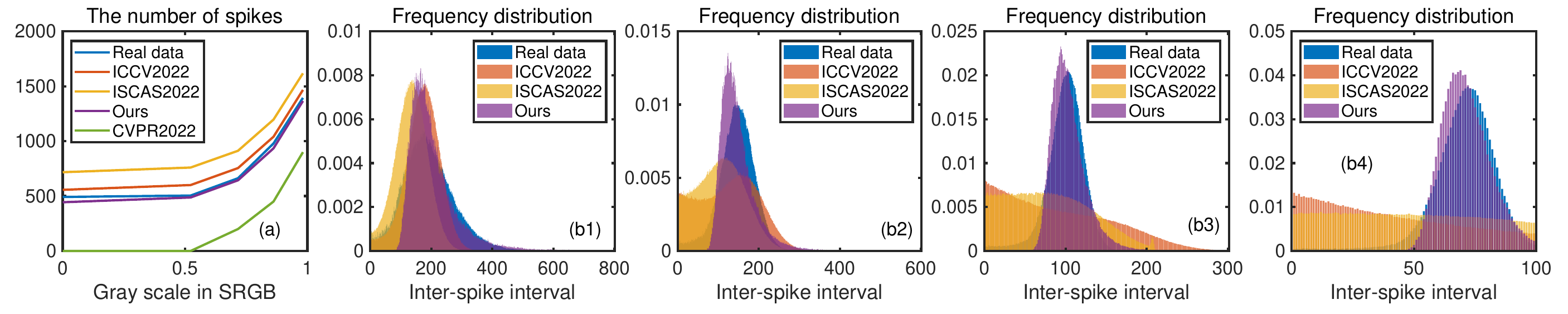}
\centering
\caption{Comparison of spike streams from different methods. (a) The average number of spikes per sampling. (b1-b4) Histograms of Inter-spike interval (ISI) \cite{spikecamera}. The grayscale of the sampled (simulated) scenes are 0, 120, 180, and 240 respectively.}\label{sim_comp}
\end{figure*}



\section{Experiments}

We verify the reliability of SCSim from two perspectives: 1. We compare the statistics between real data and simulated data. The experiment shows that the spike streams from SCSim have more similar statistical characteristics to real spike streams. 2. We find that the dataset from SCSim can improve the reconstruction performance.


\subsection{Simulators comparison}

\begin{table}[htbp]
  \centering
  \caption{Characteristics of simulators. IS denotes image sources, IF denotes inserting frames, OF denotes optical flow, BB denotes bounding box.}
    \begin{tabular}{c|cccc}
    \toprule
    Method & NeuSpike\cite{rec4} & SpikingSIM\cite{zhao2022spikingsim}  & SPCS\cite{2022scflow} & SCSim \\
    \midrule
    Ref.  & ICCV2021 & ICASSP2022 & CVPR2022& This paper \\
    IS & IF    & IF    & Render    & Render \\
    Noise & \ding{51}     & \ding{51}     & \ding{53}    & \ding{51} \\
    Depth &  \ding{53}     & \ding{53}      &  \ding{53}     & \ding{51} \\
    OF    &   \ding{53}    & \ding{53}      &     \ding{51}  & \ding{51} \\
    BB  &   \ding{53}    &   \ding{53}    &   \ding{53}    & \ding{51} \\
    \bottomrule
    \end{tabular}%
  \label{sim_tab}%
\end{table}%

The different simulators are shown in Table~\ref{sim_tab}. Previous methods \cite{zhao2022spikingsim,rec4} only consider dark current and shot noise (red part in (10)) while we also consider voltage fluctuation, capacitor mismatch and bias voltage (blue part in (11)). Besides, we compare the statistical characteristics of real spike streams and synthetic spike streams. We capture real spike streams from backgrounds with different brightness and use different methods to synthesize spike streams based on the same simulation scenarios. As shown in Fig.~\ref{sim_comp}(a), the total number of spikes in spike streams generated by SCSim is closest to real spike streams. Fig.~\ref{sim_comp}(b1-b4) shows the distribution of inter-spike interval (ISI \cite{spikecamera}) in spike streams which can reflect the dynamic process during spike camera sampling. We can find that the distribution from \cite{zhao2022spikingsim} (orange line) and \cite{rec4} (red line) is more uniform and covers a larger range when using large grayscale as simulated scenes. The shape of our distribution is similar to real data (blue line) and they are more concentrated within a specific range.

\begin{figure}[t]
\includegraphics[width=\linewidth]{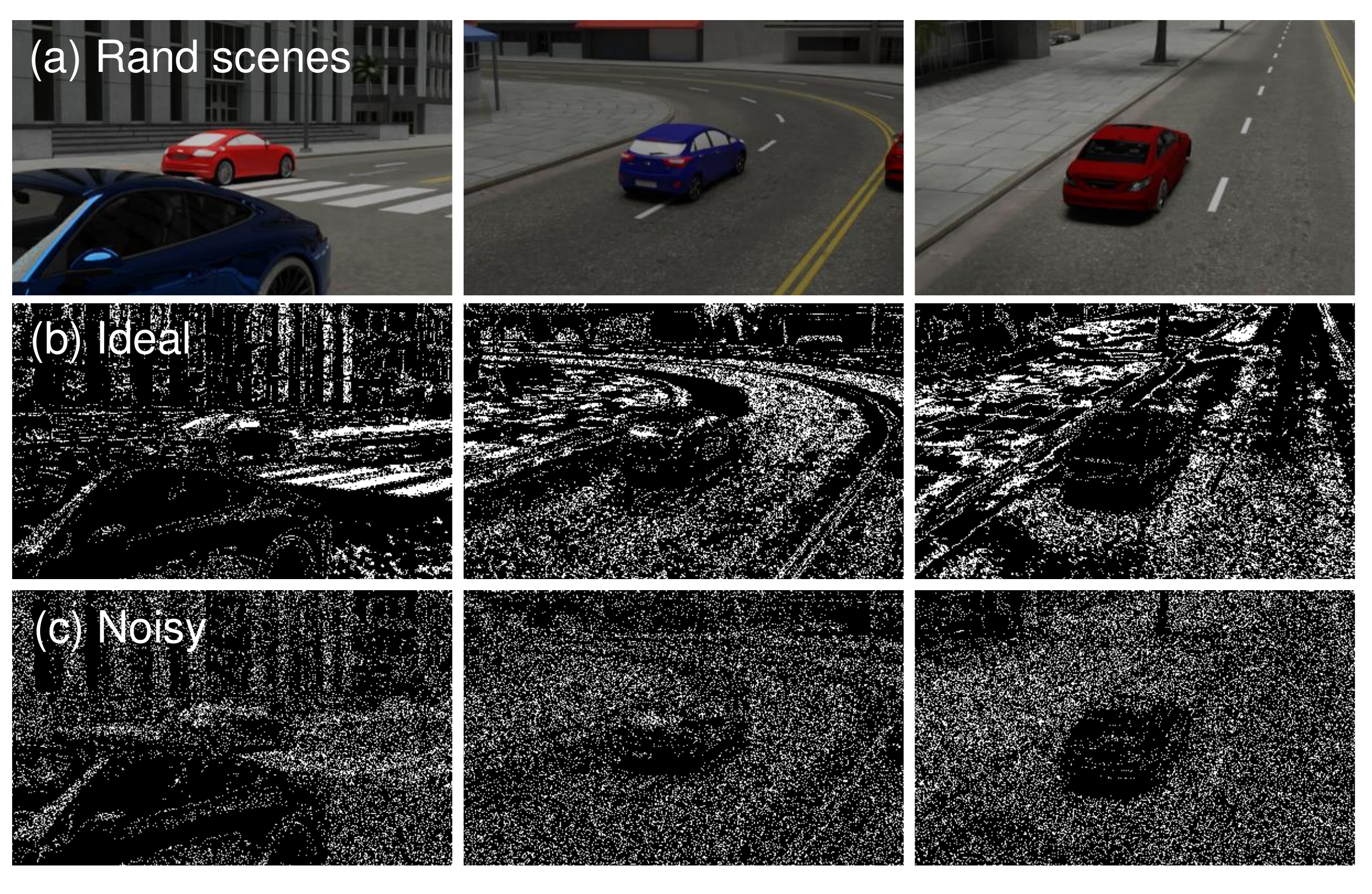}
\centering
\caption{Some examples on the RHDD dataset.}\label{dataset}
\end{figure}

\subsection{Reconstruction}

As shown in Fig.~\ref{dataset}, we use SCSim to generate the random high-speed driving dataset, RHDD, and finetune the state-of-the-art reconstructed method, WGSE.

\begin{table}[htbp]
  \centering
  \caption{Noise level of the reconstructed results on pku-spike-recon-dataset \cite{rec1}.}
    \begin{tabular}{ccccc}
    \toprule
    Method & Finetune & \cite{noise_level1} $\downarrow$ & \cite{noise_level2} $\downarrow$ & SNR $\uparrow$ \\
    \midrule
    WGSE  &   \ding{53}    & 0.316 & 0.325 & 4.304 \\
    WGSE  &   \ding{51}    & \textbf{0.273} & \textbf{0.113} & \textbf{4.314} \\
    \bottomrule
    \end{tabular}%
  \label{tab_result}%
\end{table}%

\vspace{-8pt}
\hupar{Datasets} RHDD includes 30 random driving scenarios. Each driving scene includes 500 clear images and the noisy (clean) spike stream. Real pku-spike-recon-dataset \cite{rec1} is as the test set which includes 8 scenarios.

\hupar{Train setup}  we finetune the reconstructed method, WGSE, based on RHDD. Adam optimizer is adopted to optimize the networks and the learning rate is set to 0.0001. WGSE is finetuned with a batch size of 16 on 1 NVIDIA A100 GPU.

\begin{figure}[ht]
\includegraphics[width=\linewidth]{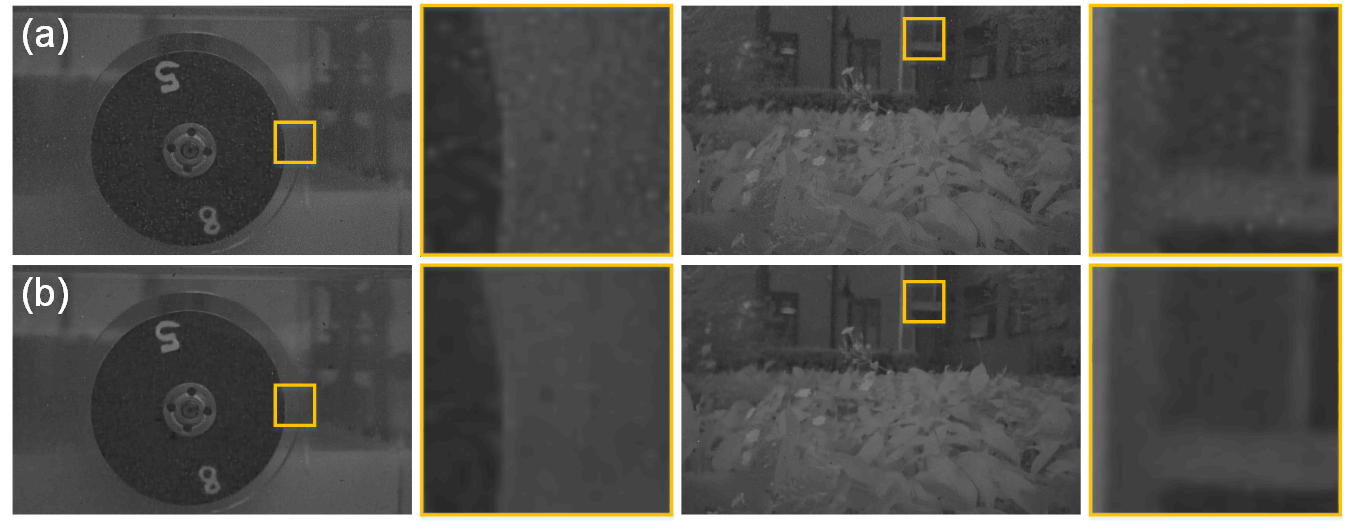}
\centering
\caption{Reconstructed results on \cite{rec1}. (a) WGSE. (b) WGSE (finetune). Please enlarge the figure for more details.
}\label{result}
\vspace{-8pt}
\end{figure}

\hupar{Results} 
we use WGSE and WGSE (finetune) respectively to reconstruct real spike streams. Due to the lack of ground truth, we use no reference metrics \cite{noise_level1}, \cite{noise_level2}, and signal to noise ratio (SNR) to evaluate the noise level of images. As shown in Table~\ref{tab_result}, WGSE (finetune) can more effectively eliminate noise in spike streams. Fig.~\ref{result} shows the visualization results. We can find that images from WGSE (finetune) are more smooth.

\section{Conclusion}

A spike camera simulator, SCSim, is proposed where we carefully model and evaluate the unique noise in the spike camera. By using designed auxiliary functions, we can easily generate random autonomous driving scenes, image sequences and visual labels. Further, image sequences are converted into spike streams based on our spike camera model . Experiment shows that SCSim can generate spike streams more accurately. Besides, SCSim can easily construct datasets and improve the state-of-the-art reconstruction method, WGSE.


\bibliography{IEEEabrv,conference_101719}
\bibliographystyle{IEEEtran}

\end{document}